\documentclass[11pt, a4paper]{deepmind}

\usepackage{booktabs}
\usepackage[numbers,compress]{natbib}

\usepackage{tocloft}
\usepackage{amssymb}
\usepackage{amsmath}
\usepackage{amsfonts}

\usepackage{xspace}

\usepackage{hyperref}

\usepackage{cleveref}

\newcommand\Tstrut{\rule{0pt}{2.6ex}}         %

\usepackage{multirow}

\newcommand{\gen}{\mathcal{G}}
\newcommand{\disc}{\mathcal{D}}

\newcommand{\myvec}[1]{\mathbf{#1}}

\newcommand{\two}{(2)}

\newcommand{\vA}{\myvec{A}}

\newcommand{\vR}{\myvec{R}}

\newcommand{\vX}{\myvec{X}}

\newcommand{\be}{\begin{equation}}
\newcommand{\ee}{\end{equation}}
\newcommand{\bea}{\begin{eqnarray}}
\newcommand{\eea}{\end{eqnarray}}
\newcommand{\beaa}{\begin{eqnarray*}}
\newcommand{\eeaa}{\end{eqnarray*}}

\DeclareMathAlphabet{\mathpzc}{OT1}{pzc}{m}{n}

\title{Large-scale multilingual audio visual dubbing}

\author[*,1]{Yi Yang}
\author[*,1]{Brendan Shillingford}
\author[*,1]{Yannis Assael}
\author[*,1]{Miaosen Wang}
\author[1]{Wendi Liu}
\author[1]{Yutian Chen}
\author[2]{Yu Zhang}
\author[1]{Eren Sezener}
\author[1]{Luis C. Cobo}
\author[1]{Misha Denil}
\author[1]{Yusuf Aytar}
\author[1]{Nando de Freitas}

\affil[*]{Equal contributions}
\affil[1]{DeepMind}
\affil[2]{Google}

\begin{abstract}

We describe a system for large-scale audiovisual translation and dubbing, which translates videos from one language to another.
The source language's speech content is transcribed to text, translated, and automatically synthesized into target language speech using the original speaker's voice.
The visual content is translated by synthesizing lip movements for the speaker to match the translated audio, creating a seamless audiovisual experience in the target language.
The audio and visual translation subsystems each contain a large-scale generic synthesis model trained on thousands of hours of data in the corresponding domain.  These generic models are fine-tuned to a specific speaker before translation, either using an auxiliary corpus of data from the target speaker, or using the video to be translated itself as the input to the fine-tuning process.
This report gives an architectural overview of the full system, as well as an in-depth discussion of the video dubbing component.  The role of the audio and text components in relation to the full system is outlined, but their design is not discussed in detail.
Translated and dubbed demo videos generated using our system can be viewed at \\
\url{https://www.youtube.com/playlist?list=PLSi232j2ZA6_1Exhof5vndzyfbxAhhEs5}.

\end{abstract}

\begin{document}

\maketitle

\renewcommand\contentsname{}
\vspace{-0.5cm}
\tableofcontents

\section{Introduction and motivation}

\begin{figure}
\centering
\includegraphics[width=\linewidth]{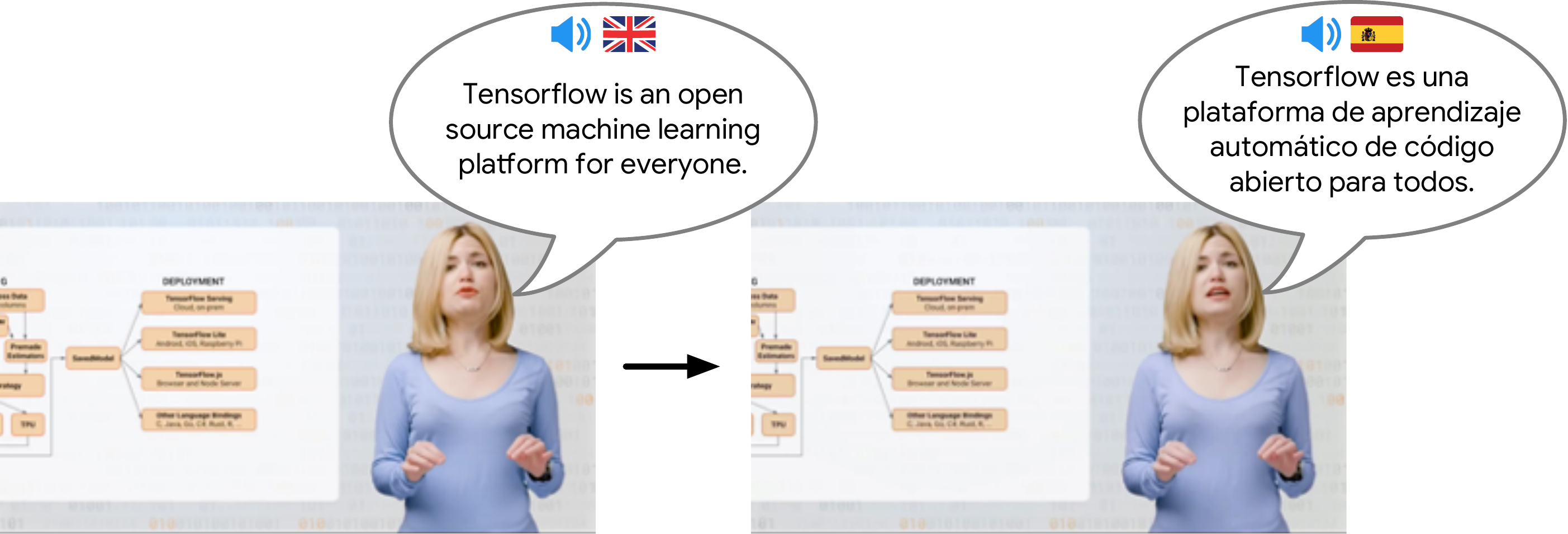}
\caption{The goal of large scale multilingual video dubbing is to translate educational videos to multiple languages with synthetic audio and synced lip movement, with the intent of creating the illusion that the videos were shot in the target language so as to increase rapport and student engagement.}
\label{fig:splash}
\end{figure}

Nearly 59\% of the Internet content is published in English~\cite{w3techs:content-language}, 
but only a quarter of its users speak English as their first language~\cite{interworld:english}.
This creates imbalances in information access and online education. Due to the language barrier, lectures on numerous timely topics are inaccessible to many people in Africa, Latin America, and other parts of the world. Automatic translation of educational videos offers an important avenue for improving online education and diversity in many fields of technology.

This work focuses on translating audiovisual content, where today more than 500 hours of videos are uploaded to the Internet per minute~\cite{statista:youtube}.
There are two major modes in audiovisual translation: subtitling and audio dubbing.
Prior studies \citep{koolstra2002pros,wissmath2009dubbing} indicate that subtitles in a foreign language decrease feelings of spatial presence, transportation, flow, and make content less accessible to beginner readers.
In contrast, audio dubbing is better for beginner readers but still results in loss of important aspects of the acting and user engagement. 
This happens because the mouth motions play a crucial role in speech understanding~\cite{mcgurk1976hearing}.

We extend audio only dubbing to include a visual dubbing component that translates the lip movements of speakers to match the phonemes of the translated audio.
This creates a more natural viewing experience in the target language, and is achieved by training a large scale visual synthesis model that generates images of lip movements conditioned on the translated audio.
The dubbing model is trained on a large scale multilingual visual speech dataset comprised of 3,700 hours of transcribed video across 20 languages.
The performance of this model is substantially improved by speaker specific fine tuning using a small amount (several minutes) of data from the target speaker.
Speaker specific data is always available from the target video itself, and when additional data is available (for example, when translating a lecture we may have access to other lectures given by the same speaker), that data can be used as well.

Our system makes use of a speaker-adaptive text to speech model similar to \citet{chen2018sample} to synthesize the translated video transcript using the voice of the original speaker.
Although we do not describe this in detail in this report, it is worthwhile to note its structural symmetry to the visual part of the system.
The text-to-speech system also employs a large scale multilingual multi-speaker model trained on thousands of hours of data that is adapted on a per-speaker basis using a small amount of additional reference speech (which can again be drawn from the target video, or from additional recordings of the same speaker).

Transcription of the target video in the source language and translation of the transcript into the target language can also be automated; however, the viewing experience in the target language depends strongly on the quality of the translated transcript.
Use of idioms or technical terms in the source language can lead to awkward phrasing in the target language, and errors in interpreting the translation can be quite jarring (for example the $y$ in $y=mx + b$ being read as a word in Spanish).
In practice, using human editors substantially improves the result over purely automated translation.

\section{System overview}

\begin{figure*}[tb]
\centering
\includegraphics[width=0.75\linewidth]{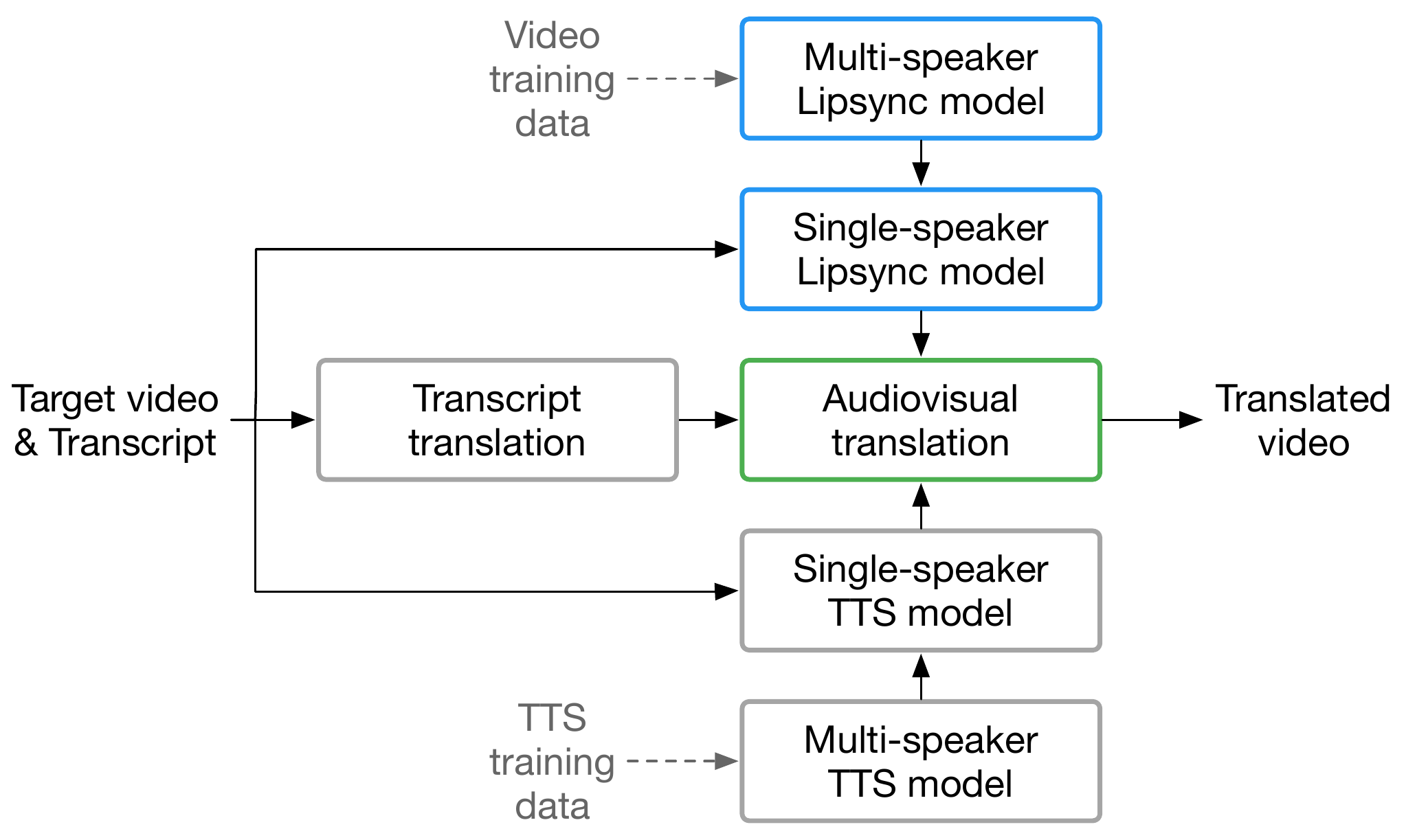}
\caption{Overview of our audiovideo dubbing pipeline.}%
\label{fig:pipeline-main}
\end{figure*}

Figure~\ref{fig:pipeline-main} gives a high level overview of our audiovisual dubbing system. Given a target video to be translated, its content is first transcribed to text and the transcription is then translated into the target language. A TTS model which is trained specifically for the speaker is used to generate the speech audio. The translated audio is then combined with the original video frames. However, the translated audio often has a different duration from the original, meaning that the video and audio speed need to be adjusted to match in length. The combined video is then fed into the single-speaker lipsync model to obtain predicted mouth frames. Finally, additional data processing steps blend the predictions with the original frames to generate the full video.

\paragraph{Lipsync model architecture}
Our Lipsync model is a large residual U-Net with three independent encoders to transform its inputs into three latent variables and a decoder to generate outputs from the latent variables. The inputs consist of video frames taken from the source video with the speaker's mouth region masked out, a set of reference frames randomly sampled from the same video to demonstrate the skin texture of the speaker, and the corresponding audio (MFCC or log mel spectrogram) aligned with the masked frames. With these inputs, the model is trained to reconstruct the masked region of the video frames.

\paragraph{Lipsync model training}
Using this architecture, we first train a multi-speaker model using the large scale dataset which consists of a wide variety of speakers and languages. Then using the trained multi-speaker model as the base, we further fine-tune a single-speaker model for each of our target speakers. 

To train the models, we follow the data collection pipeline as described in \cref{sec:lipsync:multi:data} to obtain audiovisual utterance pairs. The pipeline takes the raw videos, performs a series of data processing steps to segment the videos into utterances, extracts the face region, and performs rotation and zooming on the extracted face so that the model inputs are aligned by eye distance.

\paragraph{Inference and Rendering}

As described in \cref{sec:lipsync:multi:data}, our model's inputs and outputs are 256x256 images extracted from the original video frames that are aligned to the same orientation and eye distance as in passport photos. The predicted lips are transformed back to the original position in the full frame using an affine transform. We use Gaussian blending to put the predicted image back into the frame in a natural manner. Finally, we combine the blended frames with the translated TTS audio to yield the final video.

The following sections describe these components in more detail.
\section{Lipsync model}
\label{sec:lipsync}

We use a two-stage model for lip sync, where the first stage trains a single network over a large-scale multi-speaker multi-lingual dataset that captures the fundamental visual representation and relationship between audio and video, with the second stage further focusing on a single speaker, refining the generated visual appearance and sharpness.

Compared to existing lipsync methods that learn only on single speaker videos, our pretrained model sees more examples from different languages and different appearances, making it more robust for inference with different real and TTS audios.

Section overview:
\begin{itemize}
    \item In \cref{sec:lipsync:multi}, we describe the components necessary to construct a general multi-speaker multilingual lipsync model that generates synced face video given a speech audio signal and some contextual information.
    \item In \cref{sec:lipsync:single}, we describe how to specialize this general multi-speaker model to a single speaker, further improving its video quality, for the purposes of applying the model to a video we wish to translate and dub.
\end{itemize}

\subsection{Multi-speaker multilingual Lipsync}
\label{sec:lipsync:multi}

To train the multi-speaker multilingual model, we obtain a large collection of synced audio-video pairs from YouTube with different languages.
We first preprocess the videos, detect talking faces, and split the long videos into small utterances that are suitable for data loading.
We then annotate each utterance with face landmarks and languages, with face alignment and cropping from each frame.
We then train the single model that generates synced face videos given a speech audio waveform.

An important aspect in model training is how to prevent information leak, wherein the model learns to cheat by generating lip pixels from useless context rather than audio. We find this is crucial for generating realistic synced videos given different audios during inference. We will discuss this in greater detail in the problem formulation.

\subsubsection{Data collection}
\label{sec:lipsync:multi:data}

\begin{figure}
    \centering
    \includegraphics[width=\linewidth]{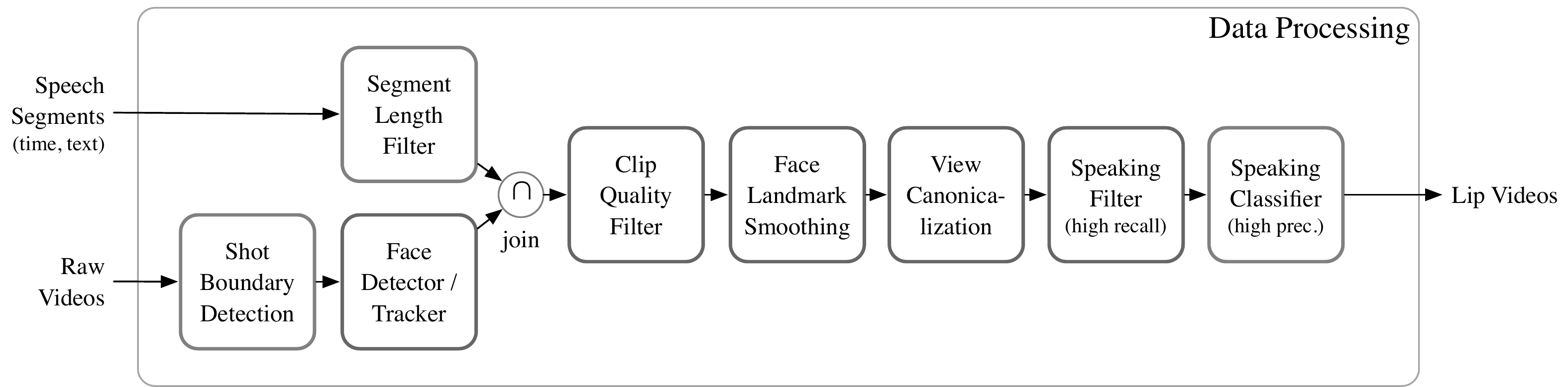}
    \caption{Data collection and filtering pipeline.}%
    \label{fig:data-collection}
\end{figure}

The multi-speaker multilingual lipsync model is trained on public videos from YouTube.
We built a large scale parallel pipeline to collect, filter, and process the videos into a format suitable for training models. Specifically, we want pairs of speech audio and sequences of face frames corresponding to the words being spoken. 
Our pipeline builds on the works of \citet{liao2013large}, which describes a procedure for generating a large automatic speech recognition dataset from verifying user transcripts on YouTube, and \citet{shillingford2018large}, which describes how to turn this into a dataset with speech audio, transcript, and face video triples. %

Starting from videos with user-generated captions, as in \citet{liao2013large}, we treat each subtitle in user-submitted captions as an utterance, and verify its audio matches the subtitle text using force alignment with an ASR model (one model for each language), following the procedure described in the aforementioned work. This yields approximately 90,000 hours of speech segments across $18$ different languages.

\begin{table}[t]
    \scriptsize
    \centering
    \begin{tabular}{c|c|c|c|c|c}
        Dataset & Year & \# Utterance & \# Hours & \# Vocab & \# Identity \\ \hline
        GRID \cite{cooke2006audio} & 2006 & 34000 & 43 & 51 & 34 \\
        LDC \cite{richie2009audiovisual} & 2009 & 1316 & 8.3 & 238 & 14 \\        
        TCD-TIMIT \cite{harte2015tcd} & 2015 & & 10.98 & 6019 & 62 \\ 
        LRW \cite{chung2016lip} & 2016 & 1000 & 167.6 & 500 & 1000 \\        
        Obama \cite{suwajanakorn2017synthesizing} & 2017 & & 17 & & 1 \\
        VoxCeleb \cite{nagrani2017voxceleb} & 2017 & 153516 & 352 & & 1,251 \\
        VoxCeleb2 \cite{chung2018voxceleb2} & 2018 & 1M & 2442 & & 6,112 \\
        Youtubers \cite{duarte2019wav2pix} & 2019 & & 46.9 & & 62 \\
        \bf Ours & 2020 & 3M & 3130 & & *464K \\
    \end{tabular}
    \caption{A comparison of our dataset to existing audio-visual video dubbing datasets. The identity count for MLVD is a rough estimate that assumes all utterances from the same video are spoken by the same person.}
    \label{tab:dataset}
\end{table}

\paragraph{Face detection and landmark smoothing}
Given a video and a corresponding speech segment, which consists of a subtitle and its start and end time, we extract that clip and run a face tracker and landmarker on the clip. If any face track does not span the entire clip, we discard it. If none remain after this filtering, we terminate early.
Furthermore, to decrease noise and jitter, we smooth the landmarks present at each frame using a Gaussian kernel over time. As the landmarks will later be used to compute crops, smoothness is important.

\paragraph{Clip quality filter}
We additionally apply a model that estimates how blurry each face is, dropping tracks that are estimated to be blurry. We also use the standard deviation of the image Laplacian as an additional clearness metric upon which we filter; this metric originally used for filtering unfocused brightfield microscopy images~\citep{pech2000diatom}.
We ignore face tracks that contain a face with an eye-to-eye distance of less than $80$ pixels, and clips with frame rates lower than $23$fps. If the frame rate exceeds $30$fps (i.e.\ $50$ or $60$fps), every second frame is dropped to give a video with a frame rate between $23$ to $30$fps.
By filtering out blurry videos and ensuring the faces are large enough (i.e.\ are high resolution), we ensure we have relatively clear and high-resolution training data.

\begin{figure}
\center
\includegraphics[width=0.75\textwidth]{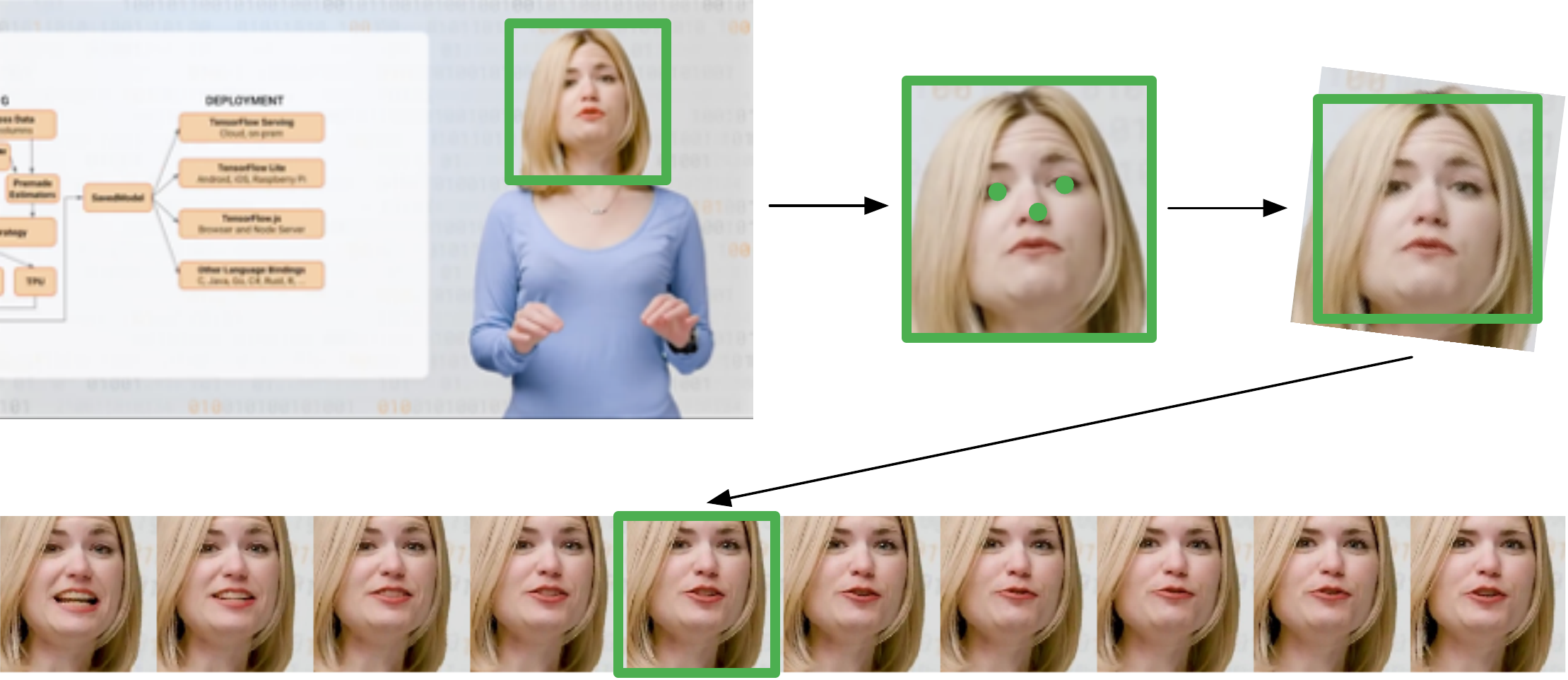}
\caption{Data processing pipeline.}
\label{fig:crop}
\end{figure}

\paragraph{View canonicalization}
Once we have a set of face tracks that span the full video clip corresponding to the speech segment, we proceed to extract per-frame facial crops, like those shown in \cref{fig:crop}. %
Using landmarks at each eye, the midpoint between the eyes, and top of the nose, paired with corresponding points on a face template in which the coordinates of these landmarks are fixed, we estimate a least-squares affine transform from the frame coordinate system to the face template coordinate system. The face template coordinate system is our crop's coordinate system.
We then orthogonalize the affine transform to remove skew, only keep scaling and rotation, yielding a Procrustes transformation (an affine transform without skew).
We use this transform to bicubicly resample the image to produce in the desired face crop.

\paragraph{Speech-mouth synchronization filter}
To eliminate data where the facial movements are not consistent with the audio, such as voice-overs, dubbed videos, cartoons, or simply other non-speaking people in the frame, we apply an active speaker detection model that checks if a sequence of face frames are consistent with a speech audio signal \citep{roth2020ava}. We use an aggressive confidence threshold to err on the side of high precision at the expense of recall, since data quality is more important than data quantity.

\paragraph{Track selection filter}
Finally, if there are multiple face tracks remaining, we keep the one where the aforementioned speech-mouth synchronization score is the highest.

\paragraph{Result}
After all of these filtering steps, approximately 5\% of the original speech segments remain.
The resulting dataset contains 3,700 hours of utterances across 20 languages.
Approximately $75\%$ of the utterances are English.
The other languages represented, in decreasing order of total duration, are Russian, Spanish, French, German, Italian, Korean, Japanese, Polish, Turkish, Dutch, Indonesian, Vietnamese, Thai, Portuguese (Brazilian), Arabic (Egypt), Romanian, Ukrainian, Hindi, and Hebrew.

Each example in the dataset is a single utterance (up to approximately 12 seconds long).
For each utterance we store the audio, the transcription, the view-canonicalized face crop, and various metadata relating the face crop to the source video (facial landmark locations, crop region, and the transformation matrix).

\subsubsection{Problem formulation}

The canonical lip synchronization problem is: given a sequence of human face images $\vX = \{X_1, \ldots, X_T\}$ and a speech audio waveform of the same length $\vA = \{A_1, \ldots, A_T\}$, we learn a model $f$ (parameterized by $\theta$) to generate a sequence of new \emph{synchronized} face images 
\begin{equation}
\vX^A = \{X^A_1, \ldots, X^A_T\} = f(\vX, \vA; \theta),
\end{equation} 
such that the mouth movements in $\vX^A$ are synchronized with $\vA$,
and the remaining part of $\vX^A$ is minimally changed compared to $\vX$ (i.e. head pose, lighting, emotion, etc.).
We say a sequence of face images $\vX^A$ is \emph{synchronized} with audio $\vA$ if a human would agree that the mouth movements of the speaker shown in $\vX^A$ are consistent with the speech in $\vA$.

To train the model $f$ supervisedly, we need the triples consisting of speech audio $\vA$; video without synchronized lips $\vX$; and $\vX^A$, the same video (in terms of background, identity, pose, etc.) but where the mouth region is synchronized to $\vA$. Unfortunately, such video is infeasible to collect. To overcome this problem, we use only synced video-audio pairs ($\vA$, $\vX^A$) to drive training.
In our model, given a sequence of images $\vX^A = \{X^A_1, \ldots, X^A_T\}$ and their corresponding original aligned audio frames $\vA =  \{A_1, \ldots, A_T\}$, we learn a model to reconstruct
\begin{equation}
\vX^A = g(\vX^A, \vA; \theta).
\end{equation}
Obviously, this is problematic as there is the trivial solution of copying the input images to the target images. To avoid that, we design a simple rectangular facial mask $M = [x_1, y_1, x_2, y_2]$ that removes all lip related information in the image, where $x_1, y_1, x_2, y_2$ represents the left, top, right, and bottom of the mask (in face crop coordinates). The face images after removing the masked region are shown in \cref{fig:arch}, and is represented as $\vX^A_M$. Hence our model is learning 
\begin{equation}
\vX^A = g(\vX^A_M, \vA, \theta). 
\end{equation}

The mask $M$ prevents the model from copying input pixels from the mouth region. However, it also removes important speaker-specific context information such as fine-grained texture of the lips and teeth. 
To compensate for this missing texture information, we further add a set of reference images $\vR_t = \{R_{t1}, \ldots, R_{tN}\}$ at every time step $t$. The reference images contain lip images taken from different time steps from the target frame, hence the face pose or lip shape could be quite different.

Our final model is defined as:
\begin{equation}
\vX^A = g(\vX^A_M, \vA, \vR; \theta),
\end{equation}
combining masked input frames $\vX^A_M$, aligned audio frames $\vA$ and reference frames $\vR$ to reconstruct the target frames $\vX^A$. 
During inference, we apply the learned model $g$ on frames $\vX^A$ with a different audio $\tilde{\vA}$ to generate 
\begin{equation}
\vX^{\tilde{A}} = g(\vX^A_M, \tilde{\vA}, \vR; \theta)
\end{equation}
as an approximation to the ideal lip sync model $\vX^{\tilde{A}} = f(\vX^A, \tilde{\vA}; \theta)$.

We train the multi-speaker lipsync model by having it synthesize lip movements for the training videos based on their original audio.
This setup is multilingual and multi-speaker since the dataset contains utterances from many speakers in many languages, but there is no translation involved since for each example we reconstitute lip movements in the original source language.
We use audio from the source video to resynthesize the lip movements in the source video, allowing the model to be trained in isolation from the rest of the system.

During training, our video-audio pairs ($\vX^A$, $\vA$) are obtained from the original video and completely synced. While during inference, our video-audio pairs ($\vX^A$, $\tilde{\vA}$) are no longer synced. This strategy allows us to build an end-to-end deep learning model with full supervised training.

\paragraph{Information leak}
Because of the way we formulate the training procedure, an important aspect of creating a model that generalizes to novel inputs is ensuring that it does not learn trivial training set correlations that will not be present in the test set when forming predictions.
For example, assume that the face crops of the training set were computed using the forehead and chin position. Thus, although the mouth area is masked, the frame itself contains information for the chin position, and thus the model can learn to infer the differences in mouth openness by the overall face position.
Instead, a correctly trained lip sync model $\vX^A = g(\vX^A_M, \vA, \vR; \theta)$ should derive mouth shape information solely from the speech audio $\vA$, mouth texture information solely from references $\vR$ and only use the masked input frames $\vX^A_M$ for head pose, lighting, and other contextual information unrelated to the speech $\vA$.
We refer to the presence of these trivial correlations as \emph{information leak}, but it is a type of overfitting or generalization error.

Our train time setting differs from the test time setting in that the training images contain a face where the masked mouth region, albeit omitted, is fully consistent. At test time, however, the mouth movements hidden in the masked region are not synced, so facial cues, head position, and body language are likely to be inconsistent with the target language speech.

\noindent There are at least 4 possible ways for information leakage to happen:
\begin{enumerate}
    \item Directly copying lip pixels from input frames $\vX^A$.
    \item Inferring lip shape from surrounding context pixels in the input masked frame $\vX^A_M$ instead of from audio $\vA$.
    \item Inferring lip shape from the scale and position of face crops inside $\vX^A_M$ which are dependent on landmark positions including those that correlate with audio $\vA$.
    \item Directly copying lip pixels from reference frames when the target frame $\vX^A$ is one of the reference frames $\vR$.
\end{enumerate}
To prevent the information leak, we adopt the following strategies:
\begin{enumerate}
    \item Use a mask $M$ to remove the mouth region and the supporting context region in each input frame.
    \item Pick a large enough mask $M$ such that the trained model cannot guess the lip shape from the surrounding pixel context in the input frames. We notice this step is a key towards reaching satisfying lip sync results during inference.
    \item Align the face images $\vX$ with only eye and nose landmarks. This means the position of the lip and chin landmarks will not influence the position of the crop. We found that not doing that leads to severe overfitting of the model, likely due to the model's size and ability to learn to predict the mouth position using the jitter in the crop position that arises as a result of using the lip and chin landmarks for computing the crop. When instead only using the eye and nose landmarks, this correlation does not arise. However, it is still important to use sufficiently many landmarks so that uncorrelated noise in the landmarks can average out, as using too few can result in a noisy Procrustes transform.
    \item Exclude the target frame from the reference frames $\vR$. To further avoid additional pose correlations, we remove the surrounding pixel context region in the reference frames, leaving only lip pixel region in the reference frames. This guarantees there is no information leak from references.
\end{enumerate}

These processing details are rarely discussed in the literature, but our experience has shown that they are absolutely critical to obtaining reliably good performance across a broad spectrum of speakers and settings.

\subsubsection{Model architecture}
\label{sec:lipsync:multi:model}

\begin{figure}[tb]
\centering
\includegraphics[width=0.8\linewidth]{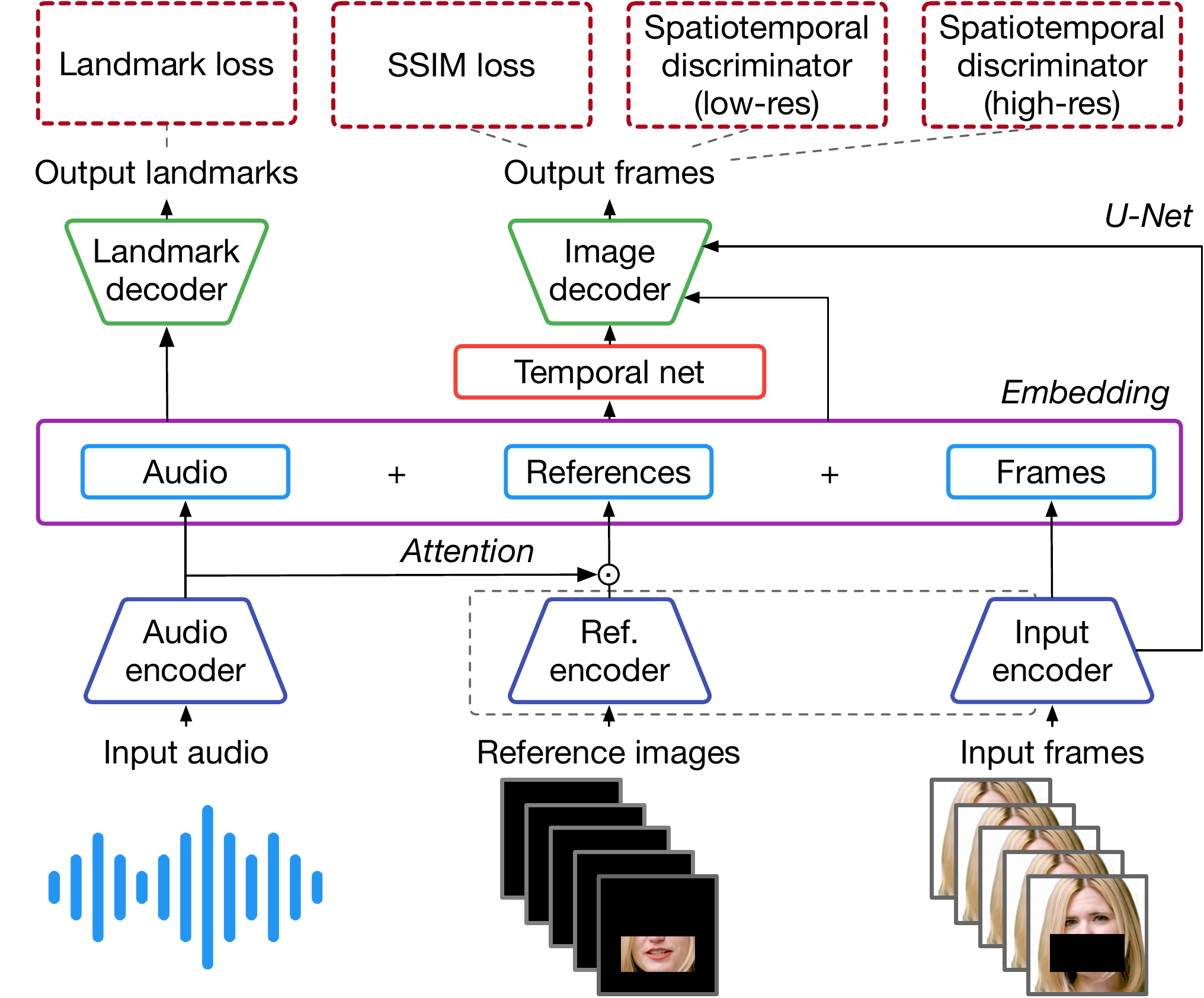}
\caption{The model takes as input the source video frames, the target audio, and reference frames that provide details for the speaker. The input frames have the mouth area masked, while the inverse mask is applied to the reference frames. The inputs and the video's language are combined to a joint embedding, which is then passed to a 1D convolutional temporal aggregation network.  Finally, the image decoder uses that embedding and the UNet connections to generate the output frames. The network is trained using a landmark regression loss on the embedding, and an SSIM and a dual discriminator loss on the output frames.}
\label{fig:arch}
\end{figure}

The lipsync network architecture is shown in \Cref{fig:arch}.
The main structure is a Residual U-Net encoder-decoder architecture that consumes the masked input frames, the source audio, and the reference frames.
Different encoders embed each input modality and the results are concatenated to form an embedding vector of size 512 for each frame of the input sequence.
These embeddings are passed to a temporal aggregation network that applies a 1d convolution to aggregate information from multiple frames across time.
The temporally aggregated features at each timestep are passed to an image decoder network whose structure is the transpose of the encoder used for the input frames.
Skip connections connect intermediate layers of the encoders to intermediate layers of the image decoder, in the typical U-Net fashion.
More specifically, the intermediate embeddings of input encoder and reference encoder are passed as the input to the intermediate layers of decoder reversely, while the embedding of audio encoder is upsampled to the same resolution of the intermediate layers to concatenate with the other embeddings. In the experiments, we find both residual blocks and skip connections are crucial to the final performance.

\paragraph{Input encoder}
The input encoder encodes the $256\times256$ resolution masked input frames $X_M$ using 9 ResNet-v2 blocks, with 2 Norm-Conv-ReLU layers per block with stride 2 for down-sampling, producing an embedding of size $1
\times 1 \times 512$.
Besides the final layer output, we also store the intermediate layer outputs from every blocks for the skip connections to the decoder.

\paragraph{Audio encoder.} The input audio $A$ is represented as a mel-spectrogram with $80$ filter banks and is encoded using 2 temporal residual blocks with $512$ filters and a kernel size of $5$, producing an embedding of size $512$.
After that, we apply a multi-layer perceptron with residual structure, making it to a list of features with same sizes as the intermediate outputs of input encoder for the skip connection in the decoder.

\paragraph{Reference encoder}
The reference encoder uses the same architecture as the input encoder.
The same set of reference frames $R=\{R_1, \ldots, R_N\}$ is used to support every frame of the input sequence.
We use the reference encoder to obtain an embedding for each reference frame independently, producing the embedding with size $N \times 512$.
For each input frame we apply soft attention over the reference frame embeddings using the audio embedding as the key.
\begin{equation}
\alpha_n = \frac{e^{-\sigma(A \cdot R_n)}}{\sum_{i=1}^N e^{-\sigma(A \cdot R_i)}}, \ R = \sum_{n=1}^N \alpha_n R_n.
\end{equation}
The soft attention allows us to form an input frame dependent convex combination of the reference embeddings, and this combination is used when synthesizing the output frame.

\paragraph{Temporal network} After combining the embeddings produced from the input frame, input audio, and reference frames in a $512$ dimensional embedding, we use a 2-layer temporal CNN with a kernel size of $3$ and channel 512, to aggregate temporal information. This architecture improves the smoothness of the generated frames.

\paragraph{Image decoder}
The decoder follows a similar architecture to the encoder in the reverse order with an image upsampling operation after each Residual block. In the U-Net architecture, the output features of each block are concatenated with the skip connection from relevant layers in the input encoder, reference encoder, and the additional upsampled audio embeddings.

\paragraph{Landmark decoder}
Besides the main structure, we also add a landmark decoder that outputs a set of 2D lip related landmark locations from the concatenated embedding that further encourage the focus on modeling shape of lips. This is modeled with an extra residual layer.

\subsubsection{Constructing model inputs and outputs}
\label{sec:lipsync:multi:preproc}

Given a batch of audio-video utterances with variable lengths, we preprocess it to suit for our model training and inference.

\paragraph{Input frames, audio, and target frames}
Given a batch of audio-video pairs with different lengths, we sample a fixed length subsequence (9 frames) from each utterance uniformly at random.
In these video subsequences, we further hide the lip region with a rectangular mask, making it as the final input frame to the model. 
After masking, the model can only see talking-unrelated pixel regions such as forehead, eyes and nose. These cannot be further removed because they provide useful context such as skin texture and specific lighting condition for each frame.

For each frame in a subsequence, we extract the corresponding audio from the time-aligned waveform.
Since audio sampling rate (16000Hz) is much faster than video sampling rate (30Hz), we extract audio features such as log-melspectrogram or MFCCs with a sampling rate of 100Hz, and then select a small sequence (24 audio frames) of audio features centered at the target frame.
Note that during inference, the audio is from another source. 
For supervised training, the current frame is also the target frame. Our model will learn to generate the missing lip region to match the target frame.

\paragraph{Frames and audio for sync discriminator}
Besides the main data inputs and outputs, for the sync discriminator, we also sample negative audio-video pairs from the utterances. The negative pairs are mixed audio and video from two different subsequences.

\paragraph{Reference frame selection}
Using the same image encoding network architecture we select and encode $10$ reference frames from the input video, this time masking the content outside the mouth. This step provides the model with the important identity information of the speaker. Hence, the diversity of the samples and the relevance to the current frame are crucial. We optimize the diversity of the references by selecting frames with K-means of $10$ clusters, using features of the landmark positions and head rotation. The relevance to the current frame is addressed using soft-attention over the reference frame embeddings with the current audio input as the key.

Reference frames are one of the key components towards generating realistic frames. During the design of reference selection algorithms, we mainly consider 3 aspects: (1) Do they capture different face varieties including pose and lighting condition? (2) Do they capture different lip shapes? (3) Could they cause information leak?

The simplest reference selection algorithm used by the literature would be selecting the first frame in each utterance.
However, selecting the first frame would result to information leakage that would not allow us to alter that frame during inference, in the setting where the target language audio does not match the pauses of the original audio. Other alternatives, include uniform sampling or sampling at equal intervals. We evaluated all of these algorithms. In the end, to avoid possible correlations and in order to provide the most diverse input to our model, we utilized a K-Means algorithm based on the 2D landmarks to cluster frames in each utterance and select the K frames that are closest to each of the cluster centers as the representative references. We find that this gives the best performance over a variety of face poses and talking styles.

\subsubsection{Model training}
\label{sec:lipsync:multi:training}

The training objective for the multi-speaker lipsync model is a combination of four terms.
Three of the terms focus directly on the quality of the generated lip images, and an auxiliary loss term focuses on ensuring that lip movements are represented accurately in the embedding space.

We only care that the model generates good quality images in the interior of the face, rather than using model capacity and training time to accurately generate the background, neck, hair, and other unchanged aspects. Therefore, it would make sense to mask the loss so that it is only applied over the face. Note that unlike the process we saw for masking the input, a masked loss cannot cause an information leak.
However, since it is difficult to mask non-trivial losses such as SSIM and the discriminator, before feeding the generated frame into the loss, we first differentiably combine the generated interior face region with the non-face region of the original frame using a mask. This also ensures the loss is applied to the edges, ensuring the model generates a smooth transition at the edge.

\begin{figure}[t]
  \centering
  \includegraphics[width=0.33\linewidth]{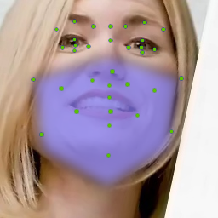}
  \caption{Polygonal mask used for processing the model output before computing the loss. A similar mask is used for stitching the face back into the original frame later.}
  \label{fig:polymask}
\end{figure}

While a generous rectangular mask is sufficient, we instead use a polygonal mask around the face (shown in \cref{fig:polymask}) computed on-the-fly from landmarks, slightly larger than the one used later for rendering. The polygonal mask is generated by computing the convex hull of the set of points given by the ears, halfway up the nose (halfway between two landmarks), the nose tip, as well as the left, right, and center of the chin. The three chin landmarks are shifted downwards to allow for the mouth to open wider than the original frame. The nose tip and convex hull is needed to ensure we produce a convex polygon regardless of the angle (e.g. profile). The polygon is then filled to produce the mask.
Compared to rectangular masking, polygonal masking reaches the same validation performance about $2\times$ faster.

\paragraph{MS-SSIM}
The Structural Similarity Index (SSIM)~\citep{wang2004image} has been widely used as a perceptual similarity metric. %
In our setup, the SSIM score is used as a differentiable loss function in multiple scales through a process of multiple stages of sub-sampling (MS-SSIM), as it can match the luminance, contrast, and structural information better than standard regression loss (L1 or L2).
However, recent literature has shown improved performance using a mix of L1 and MS-SSIM~\citep{zhao2016loss}.
\begin{equation}
    \mathcal{L}_{\text{Rec}} = \alpha \mathcal{L}_{\text{MS-SSIM}} + (1-\alpha) \mathcal{L}_{L1},
\end{equation}
where $a=0.86$. The reconstruction loss $\mathcal{L}_{rec}$ is applied only in the mouth crop area. %
The reconstruction loss is SSIM loss:
\begin{equation}
 \text{SSIM}(x, y) = \frac{\left(2\mu_x\mu_y + C_1\right)\left(2\sigma_{xy} + C_2 \right)}{\left(\mu_x^2 + \mu_y^2 + C_1\right)\left(\sigma_x^2 + \sigma_y^2 + C_2\right)}
\end{equation}
We use the MS-SSIM for training and SSIM for evaluation.

\paragraph{Spatiotemporal discriminators}
Finally, we introduce a dual discriminator setup $\disc$ to improve the quality of the generated images $\gen$, and the motion naturalness, since the reconstruction loss treats the outputs as independent frames.
Our setup is inspired by DVD-GAN~\citep{clark2019efficient} which uses a spatial and a spatiotemproral discriminator to handle image sharpness and motion, respectively. This architecture is used to efficiently handle the memory limitations of modern GPUs while maintaining a reasonable batch size. More specifically, instead of processing the whole video in high resolution, 3 random frames are selected for the spatial discriminator, for discriminating image quality, and the whole video is downsampled by $1/4$ to be processed at a lower resolution by the spatiotemporal discriminator, for discriminating motion realism.
However, this setup is not optimal for our setting. Since the network is conditioned on reference frames, using a spatial discriminator would give an incentive for the model to copy the images of the reference frames in order to trick the spatial discriminator.

To tackle this we introduce a setup with both a high-resolution image discriminator and a lower-resolution audio-visual spatiotemporal discriminator setup. More specifically, we replace the spatial discriminator that took as input 3 random frames, with a high-resolution spatiotemporal discriminator taking 3 sequential frames as input at full resolution.
The low-resolution spatiotemporal discriminator takes as input the target audio and the generated output video. To comply with modern GPU memory requirements, the video is downsampled by $1/4$.
The GAN objective is formulated using a hinge loss as described by~\citep{lim2017geometric,brock2018large}.

\begin{align}
\centering
\mathcal{L}_\disc &=  \mathop{\mathbb{E}}\limits_{x \sim data(x)}  \Big[\max(0, 1 - \disc(x))\Big]  \nonumber \\
& \qquad + \mathop{\mathbb{E}}\limits_{z \sim p(z)}
\Big[\max(0, 1 + \disc(\gen(z)))\Big],\\  %
\mathcal{L}_\gen &= \mathop{\mathbb{E}}\limits_{z \sim p(z)} \Big[\disc(\gen(z)) \Big]\\ %
\mathcal{L}_{GAN} &= \mathcal{L}_\disc - \mathcal{L}_\gen
\end{align}

The GAN loss is based on both the image discriminator and video-audio discriminator:
\begin{equation}
 L_X = \mathbb E_{X\sim P_X} [\log D_X(X)] + \mathbb E_{\hat{X}\sim P_{\hat{X}}} [\log(1 - D_X(\hat{X})]
\end{equation}
\begin{equation}
 L_{X, A} = \mathbb E_{X\sim P_X} [\log D_{X, A}(X, A)] + \mathbb E_{\hat{X}\sim P_{\hat{X}, A}} [\log(1 - D_{X, A}(\hat{X}, A)]
\end{equation}

\paragraph{Landmark reconstruction}
We introduce an auxiliary landmark loss, where we regress the Cartesian coordinates of the $13$  target landmarks around the jaw and the mouth from the embedding of each frame using an L2 loss. 
\begin{equation}
    \mathcal{L}_{\text{Land}} = \sum_{l} \left\lVert (\hat{x}_l - x_l) + (\hat{y}_l - y_l) \right\rVert^2
\end{equation}
This encourages the model to learn the locations of components of the face.

\paragraph{Training objective}
The full training objective is formed as a linear combination of the above components.
\begin{align}
\min_\theta \enspace\Big\{\alpha_{\text{Rec}} \mathcal{L}_{\text{Rec}} + \alpha_{\text{Land}}\mathcal{L}_{\text{Land}} + \alpha_{\text{GAN}}\mathcal{L}_{\text{GAN}} \big\}\enspace.
\label{eq:ms-lipsync-loss}
\end{align}

\paragraph{Optimization}
We optimize the loss in Equation~\ref{eq:ms-lipsync-loss} using a synchronous multi-worker distributed training strategy.
The loss weights were set to $\alpha_{\text{Rec}}=1$, $\alpha_{\text{Land}}=100$, and $\alpha_{GAN}=1\times10^{-4}$. Using a smaller $\alpha_{GAN}$ weight was found crucial to encapsulate the supervised objective in a GAN setup.
The models are trained using 32 GPU workers each with a batch size of 2, resulting in an effective total batch size of 64.
We use an Adam optimizer with a learning rate of $5\times10^{-4}$ for the U-Net, and $1\times10^{-4}$ for the dual discriminators.
Finally, to stabilize training, we used global norm gradient clipping of $10$ for the generator and the discriminators.

\subsection{Speaker-specific Lipsync}
\label{sec:lipsync:single}

\begin{figure*}[tb]
\centering
\includegraphics[width=\linewidth]{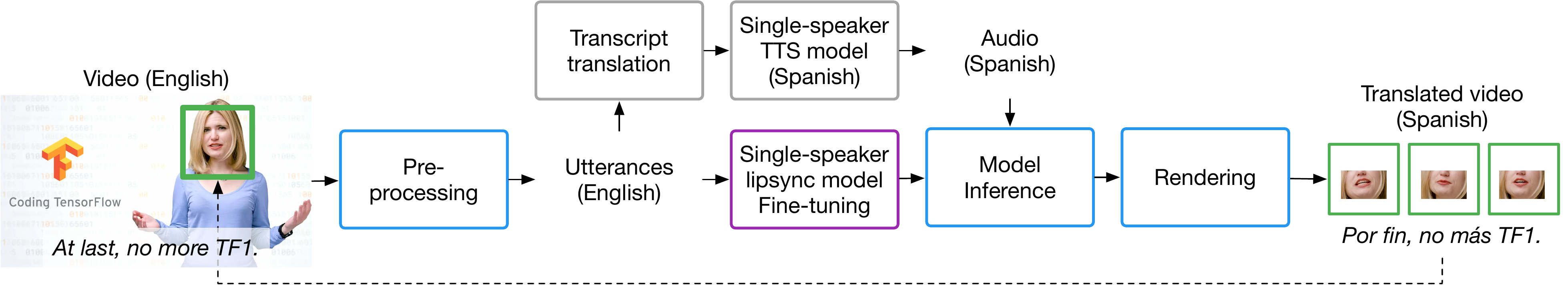}
\caption{Given a source language video, we preprocess it and generate audio-video pairs which we call utterances. The audio part is translated and used to fit a speaker-specific multilingual text-to-speech (TTS) model, which is used to produce the translated target language audio. In tandem, the video part is used to fine-tune and generate a speaker-specific video dubbing model. The model is then used to generate natural mouth motions matching the translated TTS audio, and the result is overlaid on the original video.}
\label{fig:pipeline}
\end{figure*}

In this section, we describe how we use the multi-speaker multilingual model described in the previous section for dubbing a specific video. Given a set of videos of a target speaker in the source language, we fine-tune the model on these videos. This model is later used to lipsync a target video with translated text-to-speech audio.

Instead of producing a speaker-specific model, one could simply use the multi-speaker multilingual model directly. However, to improve quality further and produce a video as similar to the source video as possible, we can fine-tune to the source video.

We assume we have, as training data, either only the source video, or the source video in addition to other videos of the same speaker. All of these videos are in the source language, and we make no assumption we have 
video of the speaker speaking the target language. We assume the target language is one of the 20 languages that the multi-speaker multilingual model was trained on.

In the following, we describe the differences of the speaker-specific training step as compared to the procedures for the multi-speaker multilingual model.

\subsubsection{Data collection}
\label{sec:lipsync:single:data}

To begin, we start with the source video, or optionally several similar such videos of the speaker in the source language.
For the sake of simplicity, we assume that only one face is speaking in a video at a time. We have found that in complex videos with rapidly alternating speech such as interviews, diarization (whether from just audio or audio-visual) is insufficiently accurate for decoding who the active speaker is. Future work is required in this direction.
For each video, we first run the face tracker and landmarker on the full video. This results in a collection of face tracks.
We stop here, as we assume the resulting face tracks all belong to the same person.
We then apply the same mouth-centered crop generating procedure as in \cref{sec:lipsync:multi:data}.

As output, we store a fully decoded copy of the video and audio extracted using the above pipeline.
All of the final face tracks are converted into the same utterance format described in the previous section, including the mouth-centered crops, corresponding audio clip, and affine transform (for mapping between frame and crop coordinates), along with additional metadata describing the audio sample indices and video frame indices at which they start and stop.
The full decoded video and additional metadata stored with the utterances ensures we have enough information to correctly render the model outputs into the correct frames.

For more complicated videos, as in the previous section, we must apply the audio-video synchronization classifier from \citet{roth2020ava}. This allows us to determine which face is speaking in a video with multiple faces. As mentioned previously, however, this breaks down easily in complicated videos with overlapping or rapidly alternating speech.

\paragraph{Chunking}
As this tracking method results in a single contiguous face track when the face is contiguously shown, unlike the \cref{sec:lipsync:multi:data} pipeline we do not start from subtitle-sized speech utterances, we need to split this long track into smaller pieces to be more easily handled by the model.

Specifically, we split each full face track into short segments that are as close in length as possible without exceeding a maximum number of frames. For a face track with $N$ frames in total, if we want at most $M$ frames in each segment, we split into at least $N / M$ segments.
To achieve that, we split each long utterance into a variable number of nearly-equal pieces, each of length not exceeding a fixed parameter.
We call each of these smaller utterances resulting from the splitting process an \emph{utterance chunk}.

\paragraph{Buffer frames} Buffer frames are extra frames added as a buffer at the beginning and end of each utterance chunk, to guarantee the generation of each frame sees enough earlier and later audio, as this contextual information can be important for predicting mouth information. In experiments, we add 10 buffer frames to the beginning and end of the chunks. When the utterance is the first one of each bigger track, there are no buffer frames, and we pad the audio with silence.

\subsubsection{Model architecture and training}
\label{sec:lipsync:single:training}

We fine-tune the pretrained model on a specific speaker for better performance.
We find this fine-tuning significantly improves sharpness and visual quality, with a smaller improvement to synchronization.

Fine-tuning follows the same training procedure as the multi-speaker model, using the aforementioned collection of single-speaker utterances instead of the large training set. This time we only train for 10,000 steps.
It is possible to also fix the encoder parameters and fine-tune the rest of the model or just the decoder, which one would imagine may make the model transfer to the target language more easily. However, in most cases, fine-tuning all model parts gives the best evaluation performance.

\section{Speech recognition, translation and voice synthesis}

In this section, we briefly explain the pipeline to produce translated audio. Specifically, we apply the following three steps in order: speech recognition to obtain the transcript, translation into the target language, speech synthesis with voice imitation.

We employ a speech recognition model to transcribe the video and split the audio track into sentences and translate the transcripts into the target language using a machine translation model. We correct errors manually in both steps of this process.

For some pairs of languages, a machine translation model may output translated text with a very different length from the source sentence. This causes problems when aligning the synthesized translated audio back to the original video. Therefore, it is common in the dubbing industry to request a human translator to produce translated sentences with as close length as possible to the original ones. This is still an open challenge for automatic machine translation in the application of dubbing.

Given translated transcripts, we would like to synthesize audio in a foreign language in the original voice of the speaker. We deploy a Tacotron 2 Text-to-Speech (TTS) model~\citep{shen2018natural} trained on a large multilingual multi-speaker speech dataset. Each speaker in the dataset is monolingual. The model receives speaker id and language id as inputs together with the sequence of text, and learns to generate speech in different languages with the same voice in the training data by switching language id input. An auxiliary loss for speaker identification can be used to improve cross-lingual generalization as demonstrated in \citet{zhang2019learning}. In order to generate speech with the same voice as the video, we apply the few-shot voice imitation method in \cite{chen2019seatts} by fine-tuning the multi-speaker TTS model on the pairs of utterances and transcripts obtained in the speech recognition step, and then synthesizing the utterances with the translated text. The quality of voice imitation depends on the quality and amount of speech data extracted from the original video, and typically requires 10 minutes or more data for good quality in naturalness and voice similarity.

Finally, we align the synthesized audio with the original video in time. If a synthesized utterance is longer than the corresponding utterance in the original language, we adjust the audio speed slightly in a post-processing step for better alignment. When the translated utterance is significantly longer than the original one, we edit the translated script manually to better match the original duration.

\section{Inference and rendering}

Inference and rendering consists of face tracking, splitting face track and TTS audio pairs into smaller pieces similar to the lengths seen at training time, performing model inference on these split utterances, and finally blending these predicted faces back into the original frames to produce the final video.

\paragraph{Input video}
We begin with the full video with audio replaced with the target-language TTS audio. This video may have slightly different timing from the original video due to the potential for manually changing lengths of clips to accommodate the target language audio being shorter or longer.

\paragraph{Processing}
This step is the same as \cref{sec:lipsync:single:data}. We find full face tracks and split these into smaller utterance chunks with buffer frames. Unlike previously however, we do not have the option of running the audio-video synchronization model to diarize multiple faces appearing in each frame, since the audio does not match.

\paragraph{Model inference}
For each utterance chunk output from the data pipeline, we apply the speaker-specific model. This generates a sequence of replacement mouth crop frames for the utterance chunk, which we then use to replace the original utterance chunk frames (excluding the buffer frames which we discard). This allows us to later reuse the associated metadata for replacing the mouth region in the full video frames.

\paragraph{Rendering} After inference, we place the generated face back to the original video. This requires a few post-processing steps. First, we generate binary image masks over the face to copy face pixels only to the face crop box and leaving the background as is. The face mask is a filled polygon with vertices computed as the convex hull of a set of points computed from facial landmarks, similar to the one shown in \cref{fig:polymask} except slightly smaller so as to only include the interior of the face.
This allows a convex polygonal region to be masked out consistently regardless of face pose, whether the view is from the front or in profile.
We apply Gaussian blur to the polygon mask to eliminate edge artifacts where the face crop meets the full frame, acting as a sort of antialiasing and also feathering the blended edge.

To return the generated crop back to the full video, for each frame, we compute the inverse of the affine transformation that was used to generate the original crop input to the model. Then we resample the image pixels using this inverse affine transformation as the mapping between the two coordinate systems. We also resample the mask in the same way, and alpha blend the original video with the restored face crop using this mask:
\[
(1-\text{blur}(\text{mask}))\cdot T(\text{crop}) + \text{mask} \cdot \text{orig.\ frame},
\] where $T$ bicubicly resamples the image according to the inverse affine transform associated with this frame.

\section{Related work}

Research in face synthesis conditioned on facial structure, audio, or even text, has gained significant attention in the recent years.
While early literature commonly predicts mouth shapes from audio~\citep{simons1990generation,yamamoto1998lip}, many recent approaches generate full talking heads~\citep{chung2017you,chen2018lip,zhou2019talking}.
Some other works only synthesize the mouth region in a given video \citep{suwajanakorn2017synthesizing,kumar2017obamanet,kr2019towards}.
Below we discuss related approaches to our work.

\paragraph{Talking heads} Generating talking faces or heads by conditioning on audio~\citep{chung2017you,chen2018lip,zhou2019talking,wiles2018x2face,zhu2018high,song2018talking,jamaludin2019you,chen2019hierarchical,vougioukas2019realistic,yi2020audio} or facial structure extracted from other videos (e.g.\ 3D meshes or landmarks)~\citep{zakharov2019few,thies2019neural} is widely studied in recent years. These approaches aim at generating a complete video of a person from scratch, often given just a single reference image. Our work instead focuses on generating only the mouth region for the purpose of audiovisual dubbing.

\paragraph{Lip synchronization}
Generating talking mouth videos by conditioning on audio~\citep{suwajanakorn2017synthesizing,kumar2017obamanet,kr2019towards,song2020everybody} is more applicable to tackling audiovisual dubbing. Further literature uses similar models conditioned on text~\citep{fried2019text}, videos of other speakers~\citep{kim2019neural}, and facial landmarks~\citep{jha2019cross}. Recent approaches improve the quality and sharpness of the generated clips using GANs~\citep{fried2019text,kr2019towards,song2020everybody}. Our approach improves upon these methods by extending the DVD-GAN~\citep{clark2019efficient} dual discriminator to the audio-visual setting.

\paragraph{Talking heads v.s. lip sync}
The main focus of \emph{talking heads} or \emph{talking faces} is to generate a realistic talking (2D or 3D) face crop with only audio and a reference image as model inputs. The reference frame is either fixed or taken from the first frame of each utterance. In contrast, \emph{lip sync} puts the generated face (or lips) back into the original video and requires it to look natural in the original video. The information about face pose, lighting condition, etc. needs be kept as original. Hence the original frame is an extra input to the model. However, simply placing the original face image as the model input allows the model to cheat. To avoid that, we carefully design the face crop algorithm and a facial mask that removes lip information from the input image.

\paragraph{Speaker adaptation} 
As the literature often focuses on single speaker models \citep{karras2017audio,suwajanakorn2017synthesizing,kumar2017obamanet,fried2019text,kim2019neural}, high quality multi-speaker model training remains as an elusive challenge. One common approach to improve performance is to train on a large multi-speaker dataset, and then fine-tune on a specific speaker~\citep{wiles2018x2face,zhou2019talking,zakharov2019few,yi2020audio}. However, unless the initial dataset is large and diverse enough, the benefits of fine-tuning can be limited. 
To remedy this limitation, we train a multi-speaker model using our large-scale multi-lingual dataset, and then perform speaker adaptation to further improve our results on speakers with a few minutes of content.

\paragraph{Multi-lingual AV translation}
The vast majority of the literature focuses on face generation for English content only. Only recent efforts \citep{jha2019cross,kr2019towards}, which came out contemporaneously with our work, have attempted to tackle the problem of audiovisual dubbing from English to Hindi. In our work, we aim for more systematic study of multilingual scenario. We provide a through study between models learned from each individual language as well as providing a pretraining/finetuning framework for video dubbing.

\paragraph{Datasets} Some of the most commonly used datasets in the literature are GRID~\cite{cooke2006audio}, LRW~\cite{chung2016lip}, 
VoxCeleb\cite{nagrani2017voxceleb} and 
VoxCeleb2\citep{nagrani2020voxceleb}. VoxCeleb2 is the largest publicly available dataset with 2,442 hours of speech. However, the dataset is mainly in English, and this can be problematic for multi-lingual translation as audio-visual correlations in different languages can have different distributions. To address this limitation we collected a large multilingual audio-visual corpus, which consists of 3,130 hours of video in 18 languages. We also collect rich information in the dataset including head pose, facial landmarks, phonemes, etc.

\section{Evaluation}

In this section we aim to demonstrate with quantitative metrics, the impact of a) our architectural decisions, b) multilingual datasets, and c) fine-tuning on a single speaker.
For the quantitative evaluation, we use the 
Fr\'echet Inception Distance (FID) \citep{heusel2017gans},
the Structural Similarity Index (SSIM), the Peak Signal-to-Noise Ratio (PSNR) perceptual quality metrics, and the variance of the image Laplacian (VLap) as a measure of image sharpness. %

\subsection{Model ablations}

\begin{table*}[t]
\centering
\resizebox{\textwidth}{!}{%
\begin{tabular}{lr|cc|cccc||rrrr}
\toprule
\multicolumn{2}{c|}{References} & \multicolumn{2}{c|}{Network} & \multicolumn{4}{c||}{Training Loss} & \multicolumn{4}{c}{Metrics} \\
\hline
Selection & \# & Attn & Temp & L1 & MSSSIM & Land & GAN & FID$\downarrow$ & SSIM$\uparrow$ & PSNR$\uparrow$ & VLAP$\uparrow$ \\
\hline
\hline \Tstrut
Random & 1 & & & \checkmark & & & & $129.4$ & $0.964$ & $29.85$ & $3.295\cdot 10^{-3}$  \\
Random & 10 & & & \checkmark & & & & $118.6$ & $0.966$ & $30.46$ & $3.314\cdot 10^{-3}$  \\
KMeans & 10 & & & \checkmark & & & & $101.2$ & $0.970$ & $31.24$ & $3.366\cdot 10^{-3}$  \\
KMeans & 10 & & & \checkmark & \checkmark & & & $97.2$ & $0.971$ & $31.24$ & $3.339\cdot 10^{-3}$  \\
KMeans & 10 & \checkmark & & \checkmark & \checkmark & & & $91.2$ & $0.972$ & $31.66$ & $3.381\cdot 10^{-3}$  \\
KMeans & 10 & \checkmark & \checkmark & \checkmark & \checkmark & & & $90.0$ & $0.973$ & $31.91$ & $3.364\cdot 10^{-3}$  \\
KMeans & 10 & \checkmark & \checkmark & \checkmark & \checkmark & \checkmark & & $88.9$ & $\mathbf{0.973}$ & $\mathbf{31.97}$ & $3.373\cdot 10^{-3}$ \\
KMeans & 10 & \checkmark & \checkmark & \checkmark & \checkmark & \checkmark & \checkmark & $\mathbf{78.5}$ & $0.970$ & $31.38$ & $\mathbf{3.438\cdot 10^{-3}}$ \\
\hline
\multicolumn{8}{l||}{\textit{You said that} \citep{chung2017you}} & $129.7$ & $0.962$ & $29.58$ & $3.261\cdot 10^{-3}$ \\
\multicolumn{8}{l||}{\textit{Towards Automatic Face-to-Face Translation} \citep{kr2019towards}} & $124.9$ & $0.964$ & $30.04$ & $3.297\cdot 10^{-3}$ \\
\multicolumn{8}{l||}{\textit{Realistic Speech-Driven Facial Animation with GANs} \citep{vougioukas2019realistic}} & $115.7$ & $0.961$ & $29.56$ & $3.314\cdot 10^{-3}$ \\
\bottomrule
\end{tabular}
}
\caption{Performance of various model configurations.}
\label{tbl:model-ablation}
\end{table*}

\Cref{tbl:model-ablation} demonstrates the performance impact of each carefully chosen architectural decision of our model on the dataset described in \cref{sec:lipsync:multi:data}. More specifically our ablation studies experiment with the following settings:

\paragraph{References} First, we experiment with the selection algorithm for reference images (random, k-means) and the number of images used.

\paragraph{Network} Then, we utilize two networks: a soft attention network that allows the network to attend to more relevant reference frames; and a 2-layered 1D CNN operating on the embeddings for temporal feature aggregation.

\paragraph{Training loss} Finally, we experiment with 4 different loss functions, L1 and MS-SSIM operating on the output images, an L2 landmark loss to regress the landmark coordinates from the per-frame embeddings, and a dual-discriminator GAN loss.

Our model utilizes 10 reference frames selected with K-Means, an attention and a temporal network, and the four training losses (L1, MS-SSIM, landmarks and GAN).
For a fair comparison with earlier literature and re-implemented the models from three milestone papers \cite{chung2017you,kr2019towards,vougioukas2019realistic} and trained them on our dataset.
The works were re-implemented as close as possible in our own experimental setup.
It can be seen in \cref{tbl:model-ablation} that our system outperforms these alternatives by a wide margin.

\subsection{Dataset ablations}

In \Cref{tbl:dataset-ablation} we demonstrate the impact of multilingual data in the performance of our system on a per language basis. From the total set of 20 languages  we chose a subset of 5 to evaluate our model separately and compare it with the full dataset. More specifically, the rows of the table represent models trained on a single language (or on all languages together in the row labeled multilang), while the columns show the performance of the model evaluated on different language. 
As it can be seen from the last column of the table, all single language trained models under-perform on the full dataset.

\begin{table*}[h!]
\centering
\begin{tabular}{l|l|rrrrr|r}
\toprule
\multicolumn{2}{l|}{\multirow{2}{*}{FID$\downarrow$}} & \multicolumn{6}{c}{Test}\\ \cline{3-8} 
\multicolumn{2}{l|}{} & \multicolumn{1}{l}{English} & \multicolumn{1}{l}{Spanish} & \multicolumn{1}{l}{French} & \multicolumn{1}{l}{Japanese} & \multicolumn{1}{l}{Turkish} & \multicolumn{1}{|l}{Multilang} \\ \toprule
\multirow{6}{*}{Train}
& English & $81.3$ & $88.7$ & $98.5$ & $73.3$ & $100.9$ & $86.3$ \\ %
& Spanish & $84.8$ & $83.1$ & $99.4$ & $75.2$ & $104.0$ & $86.6$ \\ %
& French & $81.3$ & $83.0$ & $84.4$ & $70.4$ & $98.2$ & $81.9$ \\ %
& Japanese & $92.6$ & $97.2$ & $108.4$ & $66.6$ & $116.0$ & $95.7$ \\ %
& Turkish & $92.4$ & $95.5$ & $104.3$ & $82.6$ & $87.4$ & $94.1$ \\ %
& Multilang & $75.9$ & $79.5$ & $89.7$ & $66.8$ & $91.6$ & $\mathbf{78.5}$ \\
 \bottomrule
\end{tabular}
\caption{Dataset ablations.}
\label{tbl:dataset-ablation}
\end{table*}

\subsection{Subjective evaluation}

We measure the naturalness and the audiovisual synchronization of the generated clips using the standard Mean Opinion Score (MOS) procedure. The subjects are asked to rate the naturalness of generated utterances on a five-point Likert Scale (1: Bad, 2: Poor, 3: Fair, 4: Good, 5: Excellent).
Beside naturalness, we also measure the synchronization of the generated clip and target audio using a three-point scale (1: Unmatched, 2: Partially Matched, 3: Matched).

\begin{table}[h!]
\centering
\begin{tabular}{cc|r|r}
\toprule
Training & Fine-tune  & Naturalness MOS & Sync.\ MOS \\
\hline
\hline \Tstrut
\checkmark & - & $2.35 \pm 0.07$ & $1.93 \pm 0.04$ \\
- & \checkmark & $2.45 \pm 0.06$ & $1.67 \pm 0.05$\\
\checkmark & \checkmark & $\mathbf{3.04} \pm 0.07$ & $\textbf{2.12} \pm 0.05$\\
\bottomrule
\end{tabular}
\caption{Human evaluated naturalness (no audio; scale 1 to 5), synchronization with audio (scale 1 to 3)}
\label{tbl:mos-scores}
\end{table}

In \Cref{tbl:mos-scores} we summarize the MOS scores of a pre-trained only model, a fine-tuned only model, and a model combining both. Comparing to the pre-trained model, the model trained exclusively on the target speaker showed better image quality, but worse audiovisual synchronization. This is because the pre-trained model can benefit from the diverse audio-visual pairs in the large training data and the target speaker only model are less likely to generate different skin textures and lighting conditions. Combining the two techniques, the model fine-tuned on top of a pre-trained model showed the best naturalness and audiovisual synchronization overall.

\section{Conclusions}

In this report we have described a large-scale system for audiovisual translation and dubbing.
The system translates both the audio and visual content of a target video, creating a seamless viewing experience in the target language.
The key challenge in audiovisual translation is to modify the lip movements of the speaker to match the translated audio.

To tackle this problem we collected a large multilingual dataset and used it to train a large multilingual multi-speaker lipsync model.
We additionally perform speaker-specific fine-tuning using data from each individual target speaker, which improves results and is always possible given a target video to be translated.
Our quantitative and qualitative evaluations justify our choices of architecture and procedure.

The primary goal in developing this system is the large scale translation of educational video content.
Our hope is that this system is a step towards broader accessibility of educational videos around the world.

\section{Broader impact}

This work contributes to efforts in democratising information, seeking to reduce language barriers in video media. This follows from the benefits accrued from text translation, and could have a widespread impact across sectors, from education to entertainment and gaming. In the near term, the impact of our work is focused on increasing accessibility of educational content, however the general nature of this technology means it could be applied in many different settings.

Improving the ability to lip sync means it could be possible to `puppet' an individual's face using a voice actor's speech, or other speech not spoken by that person, to generate deepfake content (for example, see face2face). Our work does not change the landscape of methods which enable this type of application. The ability to generate deepfake video content is already accessible, and our contribution does not increase the ability of malicious actors from generating this content. 

As the lip sync component is not released for others to use and since the overall system performs translation, it is unlikely for others to use it in harmful ways.
Additionally, consent was retrieved from the source video owners of the translated videos shown with this work,
and all video content generated via our system contains visible watermarks, so viewers are aware of any synthetic content displayed.

\section*{Acknowledgements}
We would like to acknowledge the following people for their contributions to this work:
Brett Wiltshire,
Brian Colonna,
Norman Casagrande,
Paul McCartney,
Sarah Henderson,
Scott Reed,
Sergio Gomez,
Vivek Kwatra,
Zachary Gleicher.

We would like to thank Martin Aguinis, Paige Bailey, and Laurence Moroney for their permission to use their Tensorflow tutorial videos to demonstrate our end-to-end video dubbing approach, and for allowing us to include the content in this report.

We also acknowledge the software community for developing the core set of tools that enabled this work, including Python~\cite{python}, Numpy~\cite{numpy}, Tensorflow~\cite{tensorflow}, OpenCV~\cite{opencv_library}, and Mediapipe~\cite{mediapipe}.

\bibliographystyle{abbrvunsrtnat}
\bibliography{refs}

\clearpage
\appendix

\section{Further implementation details}

We present more implementation details here.

\paragraph{Network Architecture}

\begin{figure}
\center
\includegraphics[width=0.66\textwidth]{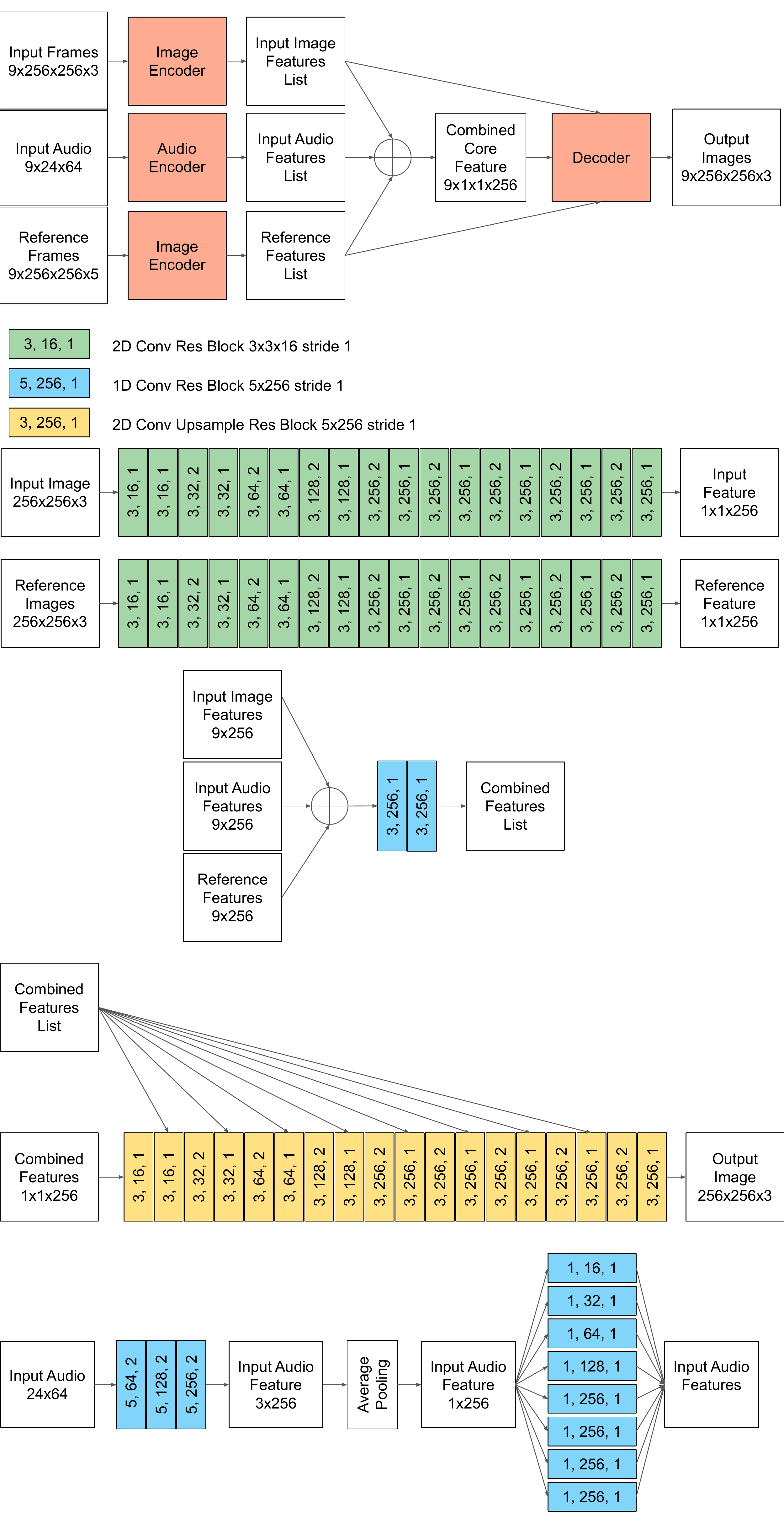}
\caption{Network architecture: the first row shows the overall architecture, and the pieces below show constituent component details.}
\label{fig:network_details}
\end{figure}

The network architecture details are shown in figure \ref{fig:network_details}. The network is consists of an image encoder, input audio encoder, reference encoder, and an image decoder, all consisting of convolutions.

\begin{table}
\label{tbl:arch_image_encoder}
\begin{center}
\begin{tabular}{llll}
    \multicolumn{1}{c}{\bf Layer}  &\multicolumn{1}{c}{\bf Size / Stride / Pad }  &\multicolumn{1}{c}{\bf Input size} &\multicolumn{1}{l}{\bf Dimension order}
\\ \hline 
    Residual Block 1 &  $16\times3\times3\times3$ / $1$ / $1$  &  $3\times256\times256$ & $C\times H\times W$\\
    Residual Block 2 &  $32\times16\times3\times3$ / $2$ / $1$  &  $16\times256\times256$ & $C\times H\times W$\\
    Residual Block 3 &  $64\times32\times3\times3$ / $2$ / $1$  &  $32\times128\times128$ & $C\times H\times W$ \\
    Residual Block 4 &  $128\times64\times3\times3$ / $2$ / $1$  &  $64\times64\times64$ & $C\times H\times W$ \\
    Residual Block 5 &  $256\times128\times3\times3$ / $2$ / $1$  &  $128\times32\times32$ & $C\times H\times W$ \\
    Residual Block 6 &  $256\times256\times3\times3$ / $2$ / $1$  &  $256\times16\times16$ & $C\times H\times W$ \\
    Residual Block 7 &  $512\times256\times3\times3$ / $2$ / $1$  &  $512\times8\times8$ & $C\times H\times W$ \\
    Residual Block 8 &  $512\times512\times3\times3$ / $2$ / $1$  &  $512\times4\times4$ & $C\times H\times W$ \\
    Residual Block 9 &  $512\times512\times3\times3$ / $2$ / $1$  &  $512\times2\times2$ & $C\times H\times W$ \\
    Output & $512\times1\times1$ & & $C\times H\times W$
\end{tabular} 
\end{center}
\caption{Lip Sync image encoder and reference encoder architecture.}
\end{table}
    
\begin{table}
\label{tbl:arch_audio_encoder}
\begin{center}
\begin{tabular}{llll}
    \multicolumn{1}{c}{\bf Layer}  &\multicolumn{1}{c}{\bf Size / Stride / Pad }  &\multicolumn{1}{c}{\bf Input size} &\multicolumn{1}{l}{\bf Dimension order}
\\ \hline
    Residual Conv1D 1 &  $128\times64\times5$ / $2$ / $2$  &  $64\times24$ & $C\times L$\\
    Residual Conv1D 2 &  $512\times128\times5$ / $2$ / $2$  &  $128\times12$ & $C\times L$ \\
    Global Pool &  $512\times6$ &  $512\times6$ & $C\times L$ \\
    Output  &  $512\times 1$ & & $C\times L$
\end{tabular} 
\end{center}
\caption{Lip Sync audio encoder architecture.}
\end{table}

\begin{table}
\label{tbl:arch}
\begin{center}
\begin{tabular}{llll}
    \multicolumn{1}{c}{\bf Layer}  &\multicolumn{1}{c}{\bf Size / Upsample / Pad }  &\multicolumn{1}{c}{\bf Input size} &\multicolumn{1}{l}{\bf Dimension order}
\\ \hline 
    Residual Block 1 &  $512\times512\times3\times3$ / $2$ / $1$  &  $512\times1\times1$ & $C\times H\times W$\\
    Residual Block 2 &  $512\times512\times3\times3$ / $2$ / $1$  &  $512\times2\times2$ & $C\times H\times W$ \\
    Residual Block 3 &  $256\times512\times3\times3$ / $2$ / $1$  &  $512\times4\times4$ & $C\times H\times W$ \\
    Residual Block 4 &  $256\times256\times3\times3$ / $2$ / $1$  &  $256\times8\times8$ & $C\times H\times W$ \\
    Residual Block 5 &  $128\times256\times3\times3$ / $2$ / $1$  &  $256\times16\times16$ & $C\times H\times W$ \\
    Residual Block 6 &  $64\times128\times3\times3$ / $2$ / $1$  &  $128\times32\times32$ & $C\times H\times W$ \\
    Residual Block 7 &  $32\times64\times3\times3$ / $2$ / $1$  &  $64\times64\times64$ & $C\times H\times W$ \\
    Residual Block 8 &  $16\times32\times3\times3$ / $2$ / $1$  &  $32\times128\times128$ & $C\times H\times W$ \\
    Residual Block 9 &  $3\times16\times3\times3$ / $2$ / $1$  &  $16\times256\times256$ & $C\times H\times W$ \\    
    Output & $3\times256\times256$ & & $C\times H\times W$
\end{tabular} 
\end{center}
\caption{Lip Sync image decoder architecture.}
\end{table}

\paragraph{Hyperparameters}

We implemented the code in Tensorflow. All network parameters are initialized from scratch with random weights. During training, we use the
Adam optimizer with learning rate 0.0005 on the generator network and 0.0001 on the discriminator network. In the dual discriminator training, we also clip the norm of the gradients of the main model and the discriminators to $10$. We train the multi-speaker model for 200,000 iterations with batch size 64 equally split on 32 GPU machines. We further finetune it on the single-speaker model for 10,000 iterations on average. 

Our model is trained fully end-to-end, with generator and discriminators learning simultaneously. Note that, there is also no data augmentation during training. The image resolution is $256\times256$. The lip mask region is $[x1=0.08, y1=0.28, x2=0.92, y2=0.95]$.

\end{document}